\pdfoutput=1

\documentclass[11pt]{article}

\usepackage{acl}

\usepackage{times}
\usepackage{latexsym}

\usepackage[T1]{fontenc}

\usepackage[utf8]{inputenc}

\usepackage{microtype}

\usepackage{inconsolata}

\usepackage{mathtools}
\usepackage{amssymb}
\usepackage{booktabs}
\usepackage{lscape}
\usepackage{multirow}
\usepackage{array,graphicx}
\usepackage[inkscapepath=svgdir]{svg}
\usepackage{flexisym}
\usepackage{tikz}
\usepackage{algorithm2e}[ruled]
\usepackage{subcaption}
\usepackage{tablefootnote}
\usepackage{dsfont}
\usepackage{xurl}
\usepackage{hyperref}
\usepackage{etoolbox}
\usepackage{enumitem}

\makeatletter
\newcommand{\removelatexerror}{\let\@latex@error\@gobble}
\makeatother

\title{Cross-lingual Human-Preference Alignment for\\ Neural Machine Translation with Direct Quality Optimization}

\author{%
  Kaden Uhlig\textsuperscript{1}, 
  Joern Wübker\textsuperscript{1}, 
  John DeNero\textsuperscript{1}, 
  Raphael Reinauer\textsuperscript{2}\thanks{Contributions made prior to joining Amazon.}\\
  \textsuperscript{1}LILT, \textsuperscript{2}Amazon\\
  \texttt{\{kaden.uhlig,joern,john\}@lilt.com}, \texttt{raphada@amazon.de}
}

\begin{document}
\maketitle
\begin{abstract}
Reinforcement Learning from Human Feedback (RLHF) and derivative techniques like Direct Preference Optimization (DPO) are task-alignment algorithms used to repurpose general, foundational models for specific tasks. We show that applying task-alignment to neural machine translation (NMT) addresses an existing task--data mismatch in NMT, leading to improvements across all languages of a multilingual model, even when task-alignment is only applied to a subset of those languages. We do so by introducing Direct Quality Optimization (DQO), a variant of DPO leveraging a pre-trained translation quality estimation model as a proxy for human preferences, and verify the improvements with both automatic metrics and through human evaluation.
\end{abstract}

\section{Introduction}

For many natural language generation (NLG) tasks, aligning models to human preferences has led to large performance gains \citep{ziegler2020rlhf}. A strong motivation for this alignment step is that much of the data on which the model was originally trained -- internet text -- is useful for language generation in general but does not match the desired output for the task. Neural machine translation (NMT) models have not involved alignment to human preferences, in part because of the assumption that supervised training data for NMT does match the desired output of the translation task. However, we show the existence of a mismatch between the NMT task and typical training data.

Machine translation is unusual among NLG tasks in that task-relevant supervised training data -- text paired with its translation -- is plentiful and publicly available. One might expect that with such a large amount of task-relevant training data, there would be no need for task-alignment. However, we identify an exhaustive list of reasons why training examples in a parallel corpus diverge from the desired output in meaningful ways (see Section~\ref{sec:train_data_mismatch}).

Machine translation is also unusual in that human preference data has been collected and published for a large number of systems, and translation quality estimation (QE) is an active research area that has benefited greatly from recent advances in large language models. We introduce a method for using quality estimation models, which themselves are trained from human preference data, in order to perform NMT task alignment. Our method, Direct Quality Optimization (DQO), is a batched online variant of Direct Preference Optimization (DPO) \citep{rafailov_direct_2023} that uses a QE model as a proxy for human preference.

We show that DQO improves translation quality in terms of BLEU, COMET22, CometKiwi22, and BLEURT, and leads to a reduction in translation errors in a human evaluation using the Multidimensional Quality Metric framework (MQM) \citep{mqmframework2014, freitag-etal-2021-experts}.

We make three notable observations when applying DQO to a multilingual model:
\begin{enumerate}[noitemsep]
    \item Task alignment increases task performance and human preference while also increasing the distance between the model's output distribution and the training data distribution.
    \item Improvements carry over to held-out languages and language families, which were not contained in the data used for DQO.
    \item Improvements in held-out languages are not limited to general behaviors required by the translation task (e.g. avoiding omissions), but include improving language-specific linguistic features not seen in the DQO alignment data, such as correctly transliterating named entities in Latvian.
\end{enumerate}

While we attribute much of the performance in held-out languages to transfer learning of general behaviors required by the translation task (e.g. avoiding omissions), the language-specific improvements in held-out languages cannot be explained by transfer learning.

Instead, these results suggest that DQO does not only increase the likelihood of the features present in its task alignment data, but also focuses the model on human preference features that it already learned during supervised training.

\section{The Task--Data Mismatch in NMT}
\subsection{Task: Human-Preferred Translations}
\label{subsec:task}
Like many NLG tasks, NMT is an open-ended problem, with multiple valid outputs for any given input, each preferred more or less by humans depending on a variety of factors, including adequacy, fluency, context, tone, style, and many other subtle features.

Because of this, the task of NMT cannot be reduced to producing valid translations, nor human-like translations, but instead requires generating human-preferred translations -- those judged as at least as good as all other valid translations.

\subsection{Training Data Mismatch}\label{sec:train_data_mismatch}
The supervised training data used in NMT comes from a variety of sources, each with notable differences from the task distribution of human-preferred translations.

\textbf{Web Data Mining}.
A large portion of parallel data is mined from massive collections of web documents, using automated methods to align source and target language segments -- e.g. the ParaCrawl \citep{banon-etal-2020-paracrawl} and CCMatrix \citep{schwenk-etal-2021-ccmatrix} datasets. This process may capture human translations, text written independently in both the source and target languages on the same topic, or the output of other MT models.
One prominent cause of task--data mismatch in automatically aligned sentence pairs is semantic misalignment. \citet{quality_glance_2022} found semantic misalignment in 15\% (ParaCrawl) and 32\% (CCMatrix) of sentence pairs in a manual quality audit.

The simplest form is complete semantic misalignment, when the source and target segments are completely unrelated. This certainly contributes to any task--data mismatch, but such pairs are easy to detect with tools such as BiCleaner \citep{bicleaner2020} or reference-free quality evaluation models such as CometKiwi22 \cite{peter-etal-2023-theres}.

Unfortunately, slight semantic misalignments of source and target are both more prevalent and much more difficult for state-of-the-art filtering systems to detect \cite{meng2024learnnoisyworldselfcorrecting}. These may include subtle yet significant differences in meaning, factual differences in numbers or names, additions and omissions, and the accompanying losses in translation adequacy.
In addition, these segments often still contain useful information that may help the model learn \cite{meng2024learnnoisyworldselfcorrecting}.

\textbf{Accidental Inclusion of Machine Translated Content}.
Web data may also include the outputs of other machine translation models, including neural, statistical and dictionary-based methods of varying quality. The impact of training on low quality machine translations is clear, however even good NMT systems' outputs differ significantly enough from natural text that classifiers can detect machine translated text with high accuracy -- and even predict which machine translation system was used to translate a given text \citep{mt_detection_2023}.

Recent research suggests that up to 57\% of translations mined from the web are multi-way parallel, meaning parallel translations of a segment can be found in more than two languages, and demonstrates a strong correlation between multi-way parallelism and low quality translations likely to be machine translated  \citep{thompson2024shocking}. The authors also found that multi-way parallel translations follow a distinct distribution, focused on low-quality content typically used for search engine optimization.

\textbf{Translator Skill Level}.
Another source of task--data mismatch in human translations is the fact that translators differ in skill level \cite{albir2017researching}. This implies that not all human translations will be equally preferred by humans.

Achieving mean human quality in translations is not the task of NMT as defined in Section~\ref{subsec:task}. We propose that neither is \textit{maximum} human quality. In theory it is conceivable that humans prefer machine-generated translations over even the best human-generated translations. Therefore, we do not want finite human skill to impose an upper limit on translation quality.

\newlength{\prevtabcolsep}
\setlength{\prevtabcolsep}{\tabcolsep}
\setlength{\tabcolsep}{0.35em}
\begin{table*}[t]
    \centering
    \begin{small}
    \begin{tabular}{llrrrrrrrr}
    \toprule
    \multirow{2}{*}{\small{Model}} & \multirow{2}{*}{\small{Lang.}} & \multicolumn{4}{c}{\small{FLORES+ devtest}} & \multicolumn{4}{c}{\small{NTREX}} \\
     &  & \small{BLEURT} & \small{COMET22} & \small{CometKiwi22} & \small{BLEU} & \small{BLEURT} & \small{COMET22} & \small{CometKiwi22} & \small{BLEU} \\
    \midrule
    Baseline & All & 0.7614 & 0.8741 & 0.8387 & 34.19 & 0.7016 & 0.8359 & 0.8099 & 30.31 \\
    DQO & All & \textbf{0.7790} & \textbf{0.8873} & \textbf{0.8508} & \textbf{35.31} & \textbf{0.7212} & \textbf{0.8525} & \textbf{0.8255} & \textbf{31.21} \\
    \midrule
    Baseline & $\mathcal{T}$ & 0.7231 & 0.8417 & 0.8272 & 34.50 & 0.6677 & 0.8040 & 0.7979 & 32.62 \\
    DQO & $\mathcal{T}$ & \textbf{0.7381} & \textbf{0.8559} & \textbf{0.8401} & \textbf{35.34} & \textbf{0.6854} & \textbf{0.8209} & \textbf{0.8137} & \textbf{33.16} \\
    \midrule
    Baseline & $\mathcal{T}^c$ & 0.7691 & 0.8805 & 0.8410 & 34.13 & 0.7084 & 0.8423 & 0.8123 & 29.85 \\
    DQO & $\mathcal{T}^c$ & \textbf{0.7872} & \textbf{0.8935} & \textbf{0.8529} & \textbf{35.30} & \textbf{0.7284} & \textbf{0.8588} & \textbf{0.8278} & \textbf{30.82} \\
    \midrule
    Baseline & $\mathcal{R} \cap \mathcal{T}^c$ & 0.7802 & 0.8820 & 0.8447 & 36.46 & 0.7202 & 0.8432 & 0.8154 & 33.01 \\
    DQO & $\mathcal{R} \cap \mathcal{T}^c$ & \textbf{0.7967} & \textbf{0.8936} & \textbf{0.8557} & \textbf{37.54} & \textbf{0.7391} & \textbf{0.8593} & \textbf{0.8307} & \textbf{34.13} \\
    \midrule
    Baseline & $\mathcal{R}^c$ & 0.7549 & 0.8787 & 0.8364 & 31.17 & 0.6934 & 0.8413 & 0.8084 & 25.84 \\
    DQO & $\mathcal{R}^c$ & \textbf{0.7751} & \textbf{0.8934} & \textbf{0.8493} & \textbf{32.46} & \textbf{0.7147} & \textbf{0.8581} & \textbf{0.8242} & \textbf{26.61} \\
    \bottomrule
    \end{tabular}

    \end{small}
    \caption{\textbf{Evaluation metrics on FLORES+ devtest and NTREX} with the NVIDIA Megatron EN-X model, before and after task-alignment using DQO. Results are shown for relevant groupings of the 30 target languages: all languages, languages used in DQO ($\mathcal{T}$), languages not used in DQO ($\mathcal{T}^c$), languages not used in DQO, but related to those used in DQO ($\mathcal{R} \cap \mathcal{T}^c$), and languages neither used nor related to the languages used in DQO ($\mathcal{R}^c$). }
    \label{tab:flores_eval}
\end{table*}

\setlength{\tabcolsep}{\prevtabcolsep}

\textbf{Translationese}.
Another common issue is a phenomenon known as \textit{translationese}, the observation that human-translated text in a given language differ in distribution from text written independently in that language. Specifically, translated text shows signs of interference from the source language's grammar, word order and word choice, as well as source-language-independent effects of the translation process itself, such as simplification and avoidance of unique language features \citep{koppel-ordan-2011-translationese, simplificationInTranslation1998, TirkkonenCondit2004UniqueItems}.
These effects are significant enough that classification models can distinguish translated and original text with high accuracy \citep{baroni2006, sominsky-wintner-2019-automatic}, as well as identifying the source language of the text \citep{koppel-ordan-2011-translationese}.
As humans show a consistent preference for translations closer to the distribution of original text rather than translationese \citep{riley-etal-2020-translationese, freitag-etal-2022-natural}, this creates an inherent task--data mismatch for training data translated in the source--target direction.

\textbf{Source--Target Domain Mismatch}.
Translation pairs in the other direction, target--source, are better aligned with human preference, as the target labels are drawn from the original text distribution rather than from translationese.
Unfortunately, they suffer from another subtle source of task--data mismatch found in human translations: Source--target domain mismatch \citep{sourceTargetDomainMismatch2021} is the observation that speakers of different languages tend to discuss different topics. For instance, a Cherokee newspaper is likely to report on different topics than an Icelandic newspaper would, and translations of these topics would remain representative of the Cherokee domain. This effect is especially pronounced for low-resource language pairs \citep{sourceTargetDomainMismatch2021}.

If one were to avoid the task--data mismatch of translationese by using only target--source translation pairs, the training data may lack key information about topics found only in the source domain. Because the task is translation from the source domain into the target language, this, too, would represent an unavoidable task--data mismatch.

\section{Human Preference Learning for LLMs}
Supervised data showing chat-based dialog between humans and AI assistants was, prior to the wide availability of such agents in the form of LLMs, understandably rare. Even with the advent of high quality proprietary and open-source models, which one could sample to create synthetic data, there is a fundamental task--data mismatch: the task is not to imitate an existing AI assistant, but (ideally) to train a new state-of-the-art model.

LLM training instead follows a two-step process:
\vspace{-.3cm}
\begin{enumerate}[noitemsep]
    \item Supervised learning on massive amounts of web data.
    \item Task alignment using instruction fine-tuning and human preference learning.
\end{enumerate}
\vspace{-.1cm}

In step one, the actual task for which the model is optimized is predicting the next token in documents taken from the web. This, when done at scale and with a variety of data sources, provides the model with extensive world knowledge and understanding of a wide array of styles and document types.

This is then followed by instruction fine-tuning, a comparatively brief round of supervised learning on human- or AI-labeled examples of dialogues, which brings the model's output distribution into the general neighborhood of desired behavior. Finally human preference learning, using actual human rankings aligns the model with the desired task: producing human-preferred responses to questions and dialog, while remaining helpful and harmless \citep{bai2022training}.

Direct Preference Optimization (DPO) is a preference learning algorithm that trains on preference pairs of the form $(x, y_w, y_l)$, with $x$ being a model input, and $y_w$ and $y_l$ being two potential model outputs for the input $x$, marked as chosen (winning) or rejected (losing) by humans during data collection \citep{rafailov_direct_2023}, using the loss function:
\begin{flalign}
& \mathcal{L}_{\text{DPO} }(x, y_w, y_l) = && \\
& \log \sigma \left(\beta \log \frac{\pi_\theta(y_w|x)}{\pi_{\text{ref}}(y_w|x)} - \beta \log \frac{\pi_\theta(y_l|x)}{\pi_{\text{ref}}(y_l|x)}\right),
\notag
\end{flalign}
where $\sigma$ is the logistic function.

\section{Direct Quality Optimization for NMT}\label{sec:DQO}

Because of its stability and ease of use, we select DPO as the basis for our experiments with human preference learning as a form of task alignment for NMT. As a proxy for human preferences, we use the CometKiwi22 quality estimation model to score and compare multiple translations of a given source \cite{rei_cometkiwi_2022}. CometKiwi22 is highly multilingual and has been shown to correlate well with human preference \cite{kocmi2024navigatingmetricsmazereconciling}. To verify that our method is not dependent on the specific choice of quality estimation model we conducted a brief experiment using MetricX \cite{juraska-etal-2023-metricx} instead of CometKiwi22 and obtained very similar results.

Our main experiments are run with the NVIDIA Megatron English--Many model\footnote{\url{https://catalog.ngc.nvidia.com/orgs/nvidia/teams/nemo/models/megatronnmt_en_any_500m}}, a 500M parameter encoder-decoder model, which supports translating from English into 30 languages\footnote{The model was originally trained to support 32 languages, but we found that translating into Arabic and Slovak resulted in degenerate output.} from 14 language families, listed in Table~\ref{tab:language_codes}. We denote the complete list of supported target languages as $\mathcal{M}$.

\begin{table}[h]
\small
\centering
\begin{tabular}{ll}
\toprule
\textbf{Language Family} & \textbf{Languages (ISO 639-1)} \\
\midrule
Baltic                   & lt, lv                             \\
Germanic                 & da, \textbf{de}, nl, no, sv                 \\
Romance                  & \textbf{es}, fr, it, pt, ro                 \\
Slavic                   & bg, cs, hr, pl, \textbf{ru}, sl, uk     \\
Uralic                   & et, fi, hu                         \\
Other                    & el, \textbf{hi}, id, ja, ko, tr, vi, \textbf{zh} \\
\bottomrule
\end{tabular}
\caption{\textbf{Target languages supported by the NVIDIA Megatron En-X model}. The category ``Other'' contains all languages that are the only supported representative of their language family. The languages on which we apply task alignment are denoted in boldface.}
\label{tab:language_codes}
\end{table}

The model's multilingual nature allows us to apply task alignment to a subset of language pairs and observe the effects on unrelated languages, with minimal risk of exposing the model to any new information in those languages.

Any improvements in those languages must either apply to all languages (such as avoiding omissions or additions), or are language specific, and can only have come from previously unused latent knowledge acquired during supervised training.

In our experiments, we selected Chinese, German, Hindi, Russian and Spanish as the target languages used during task alignment, termed $\mathcal{T}=\{de,es,hi,ru,es\}$. Let $\mathcal{T}^C = \mathcal{M} \setminus \mathcal{T}$ be the set containing the 25 target languages not represented during task alignment, $R$ be the set of languages related to at least one language in $\mathcal{T}$ (defined as belonging to the same language family), and $\mathcal{R}^C = \mathcal{M} \setminus \mathcal{R}$ be the languages unrelated to any of the languages used in task alignment. An overview of how many languages belong to each set is shown in Table~\ref{tab:language_categories}.

\begin{table}[h]
\small
\centering
\begin{tabular}{llr}
\toprule
\textbf{Subset} & \textbf{Definition} & \textbf{Size} \\
\midrule
$\mathcal{T}$ & Languages seen in DQO & 5 \\
$\mathcal{T}^C$ & Languages not seen in DQO & 25 \\
$\mathcal{R}$ & Languages related to $\mathcal{T}$ & 19 \\
$\mathcal{R}^C$ & Languages unrelated to $\mathcal{T}$ & 11 \\
\bottomrule
\end{tabular}
\caption{\textbf{Target languages supported by the NVIDIA Megatron EN-X model}, categorized by their relationship with the languages selected for task alignment.}
\label{tab:language_categories}
\end{table}

As the seed dataset from which to draw source sentences for human preference learning, we use the source side of a mixture of publicly available English--German MT datasets (see Appendix~\ref{appendix:seed_dataset}).

From this dataset, we sample 8000 source segments. For each source segment, we sample a target language from $\mathcal{T}$, the languages used for task alignment, and use the current policy model to sample 64 translations into that language using combined Top-K and Top-P sampling, with $K=40$, $P=0.8$ \cite{fan-etal-2018-hierarchical, Holtzman2020The}. We also add the greedy translation for each source segment, obtaining a total of 520\,000 translations.

Letting the output of the CometKiwi22 Quality Estimation (QE) model for a source $x$ and translation $y$ be $r_{QE}(x, y)$, we build a relation $\succ_x$ as a proxy for true human preferences:
\vspace{-.1cm}
\begin{equation*}
  y_1 \succ_x y_2 \equiv r_{QE}(x, y_1) > r_{QE}(x, y_2) + \varepsilon
\end{equation*}
where $\varepsilon \geq 0$ is a tolerance parameter to help mitigate proxy model noise. We set $\varepsilon = 0.005$.

To construct preference pairs, we then select the highest scoring translation per source segment as $y_w$ and uniformly sample a single $y_l$ from all remaining translation candidates that satisfy $y_w \succ y_l$ under our proxy model.

This results in slightly under 8000 preference pairs (occasionally the maximum difference in COMET22 score between a segment's highest and lowest scoring sampled translations is less than $\varepsilon$, in which case we do not produce a preference pair), we run DPO training with a batch size of 8192 tokens (counting source, chosen and rejected tokens), a learning rate of $1\mathrm{e}{-6}$ and $\beta = 0.5$. See Appendix~\ref{appendix:hyperparameters} for a full list of hyperparameters.

At this point, we train on the preference pairs using standard DPO for 8 epochs, after which we sample a fresh set of source segments from the seed dataset, sample translations from the policy model, create a new set of preference pairs, and begin the training again. This resampling process helps ensure that the preference pairs remain relevant to the policy model during training, and leads to substantial performance improvements. In total, we perform 5 rounds of DPO training. We call this end-to-end process Direct Quality Optimization (DQO), detailed formally in Algorithm~\ref{alg:DQO}.

\SetKwComment{Comment}{\lhd~}{}
\SetKwInput{KwHyper}{Parameters}
\SetKw{KwSample}{sample}
\SetKwFunction{AlgDPO}{DPO}
\SetKwFunction{AlgGreedy}{Greedy}
\RestyleAlgo{ruled}
\SetAlgoCaptionSeparator{\textbf{Algorithm 1:~}}

\begin{figure}[ht]
\begin{subfigure}{\columnwidth}
\removelatexerror

\begin{algorithm*}[H]
\caption{Direct Quality Optimization}
\KwHyper{preference relation $\succ$, number of rounds $n$, epochs per round $m$, epoch size $d$, learning rate $\alpha$, DPO regularization $\beta$, sampled translations per source $k$}
\KwIn{Source language seed dataset $S$, reference-free QE model $r_{QE}$, reference model $\pi_{\text{ref}}$}

$\pi_\theta \gets \pi_{\text{ref}}$\;
\For{round $i = 1, 2, \dots, n$}{
  $X \gets$ \KwSample $d$ sentences from $S$\;
  $P \gets \varnothing$\;
  \ForEach{source $x \in X$} {
    $g \gets$ \AlgGreedy\!$_{\pi_\theta}$($x$)\;
    $Y \gets$ \KwSample $k$ translations of $x$ from $\pi_\theta$\;
    $Y_{+} \gets Y \cup \{g\}$\;
    $y_w \gets \text{argmax}_{y \in Y_{+}}\, r_{QE}(x, y)$\;
    $Y_l = \{y' \in Y_{+} | y_w \succ_x y' \}$\;
    \uIf {$Y_l \neq \varnothing$} {     
      $y_l \gets$ \KwSample $y \in Y_l$\;
      $P \gets P \cup \{(x, y_w, y_l)\}$\;
    }
  }
  \For{epoch $j = 1, 2, \dots, m$}{
    $\pi_\theta \gets$ \AlgDPO{$\pi_\theta, \pi_{\text{ref}}, P, \alpha, \beta$}\;
  }
}
\end{algorithm*}
\end{subfigure}
\caption{ \textbf{Direct Quality Optimization (DQO).} \protect\AlgGreedy$\!_{\pi}(x)$ is the translation of $x$ produced with greedy search and the model $\pi$. \protect\AlgDPO refers to Direct Preference Optimization -- for full implementation details see \citet{rafailov_direct_2023}.\label{alg:DQO} }
\end{figure}

DQO can be viewed as a batched online version of DPO, as the updates are performed on batches of data sampled from the policy model.

\section{Experimental Results}
\subsection{Automatic Quality Metrics}
We evaluated the model pre- and post-task alignment on the FLORES+ \citep{nllb-22} and NTREX \citep{federmann-etal-2022-ntrex, barrault-etal-2019-findings} datasets, both of which cover all of the languages supported by the Megatron model.

We use corpus-level sacreBLEU\footnote{Signature: \texttt{nrefs:1|case:mixed|eff:no|tok:13a|} \texttt{smooth:exp|version:2.4.0}. For JA and ZH, we additionally use the \texttt{mecab-ja} and \texttt{mecab-zh} tokenizers.} \cite{post-2018-call} as well as three neural evaluation models: Reference-free CometKiwi22 \citep{rei_cometkiwi_2022}, reference-based COMET22 \citep{rei-etal-2022-comet}, and BLEURT \citep{sellam2020bleurt}.

Here it is important to note that the CometKiwi22 model was used as a proxy for human preferences in this experiment, and was thus directly optimized for. The scores from the other two neural evaluation models are thus more reliable measures of general model quality, and allow us to check for reward hacking, i.e. over-optimization for the CometKiwi22 model at the cost of performance.

\begin{figure}[ht]
    \centering
    \includesvg[width=\columnwidth]{dqo_change_in_bleurt_flores_devtest.svg}
    \caption{\textbf{Changes in BLEURT on FLORES+ devtest} with the NVIDIA Megatron EN-X model, before and after task alignment with DQO. Languages used in DQO are bolded.}
    \label{fig:flores_devtest_bleurt}
\end{figure}

Results are reported in Table~\ref{tab:flores_eval} and Figure~\ref{fig:flores_devtest_bleurt}.
We find that DPO task alignment increases all three neural quality metrics on both datasets for each of the 30 target languages.
BLEU scores increased for all languages on both datasets, with the exception of Hindi, which decreased by $0.40$ BLEU on NTREX and $0.4$ BLEU on FLORES+ devtest, despite showing improvements on the three neural metrics, like all other languages.

Significantly, translation quality, as measured by all four translation quality metrics, improved even for target languages unrelated to the languages used in DPO task alignment. See Appendix~\ref{appendix:evalmetrics} for the metrics for each individual language.

\begin{figure}[ht]
    \centering
    \includesvg[width=\columnwidth]{dqo_ablation_dpo_vs_sft_flores_dev_avg.svg}
    \caption{\textbf{Mean BLEURT-20 on FLORES+ dev at each round of DQO} with the NVIDIA Megatron EN-X model, using either Direct Quality Optimization (DQO) or Supervised Fine-Tuning (RAFT) to update the model.}
    \label{fig:dpo_vs_sft_ablation_mean}
\end{figure}

To ablate the use of DPO as the update step within DQO, we perform a comparative experiment identical to DQO as described in Section~\ref{sec:DQO} and Algorithm~\ref{alg:DQO}, but using standard supervised fine-tuning (SFT) on the preferred translation instead of DPO. Note that this is equivalent to Reward rAnked Fine-Tuning (RAFT) \cite{dong2023raft}.
Figure~\ref{fig:dpo_vs_sft_ablation_mean} shows mean performance for all language pairs through the 5 rounds of DQO. RAFT's lower performance is primarily due to catastrophic behavior for FR, JA, KO, and ZH. The poor performance of RAFT on FR, JA and KO, which were not used in RAFT training, could potentially be explained by a failure to generalize from the training data languages. Regarding ZH, which was one of the five languages used in training, we suspected unintentionally reversed labels. However, careful inspection of the training preference pairs showed no issues. See Appendix~\ref{appendix:dpo_vs_sft_ablation_flores_dev_all}, Figure~\ref{fig:dpo_vs_sft_ablation_flores_dev_all} for charts of individual language performance and Figure~\ref{fig:dpo_vs_sft_ablation_flores_dev_avg_no_outliers} for mean performance after excluding the above mentioned outliers. We leave a deeper analysis to future work.

\subsection{Training Data Perplexity}\label{sec:trainingppl_res}
In order to confirm the existence of a task--data mismatch, we examine DQO's effect on model perplexity over the training data. As we do not have access to the training data used for the NVIDIA Megatron English-Many model, we repeat the above experiment with a proprietary encoder-decoder model trained on publicly available English-to-German data using the NVIDIA NeMo framework \citep{kuchaiev2019nemotoolkitbuildingai} (See Appendix~\ref{appendix:seed_dataset}).
The model architecture is similar to the Megatron model, and follows the deep encoder, shallow decoder recipe from \cite{kasai2021deepencodershallowdecoder}, but is larger, with a model width of 2048, a feed-forward width of 8192, 21 encoder layers, 2 decoder layers, and a 32k token vocabulary, totaling 1.3B parameters.

We apply DQO to this model as with the Megatron model, however using only English--German preference pairs. After applying DQO, we see large improvements in CometKiwi22 and COMET22 for a variety of evaluation datasets, confirming that DQO worked as expected. The arithmetic mean of perplexity over a random sample of 1~million segments from the training data increased from $7.219$ (baseline model) to $9.435$ (DQO), confirming that the improvements in test data preference correspond to a reduction in the model's fit to the training corpus.

\begin{table*}[t]
\small
\centering
\begin{tabular}{p{0.07\textwidth} p{0.89\textwidth}}
    \toprule
    Source & … under the leadership of \textbf{Deng Xiaoping}. \\
    Baseline & … tika veiktas \textbf{\textit{Deng Xiaoping}} vadībā. \\
    DQO & … tika veiktas \textbf{Dena Sjaopina} vadībā. \\
    \midrule
    Source & … that \textbf{Carolyn Wilson} of the OHA had stolen their security deposits … \\
    Baseline & … ka OHA \textbf{\textit{Carolyn Wilson}} bija nozagusi viņu drošības depozītus … \\
    DQO & … ka OHA darbiniece \textbf{Karolīna Vilsona} bija nozagusi viņu drošības depozītus …\\
    \midrule
    Source & … that it was \textbf{Louis Jourdain}, 16-year old son of … \textbf{Floyd Jourdain}. \\
    Baseline & … ka tas bija \textbf{\textit{Louis Jourdain}}, 16 gadus vecs … \textbf{Floida \textit{Jourdaina}} dēls. \\
    DQO & … ka tas bija \textbf{Luiss Džordēns}, 16 gadus vecs … \textbf{Floida Džordēna} dēls. \\
    \midrule
    Source & \textbf{King Sejong} was the fourth king of the \textbf{Joseon} Dynasty … \\
    Baseline & \textbf{\textit{King Sejong}} bija ceturtais karalis no \textbf{\textit{Joseon}} dinastijas … \\
    DQO & \textbf{Karalis Sedžons} bija ceturtais \textbf{Džosona} dinastijas karalis … \\
    \bottomrule
\end{tabular}
\caption{\textbf{Examples of translations into Latvian from the FLORES+ data set before and after DQO.} Names are bolded to highlight the DQO model's increased ability to consistently transliterate names into Latvian orthography. Names that are incorrectly transliterated are in italics. Sentences are truncated to avoid dataset leakage.}
\label{tab:latvian}
\end{table*}

\subsection{Discussion}
The nearly-universal improvements for both FLORES+ and NTREX in all four automatic translation quality metrics (Table~\ref{tab:flores_eval}) provide strong evidence that DQO is a suitable task-alignment algorithm for the task of producing human-preferred translations.

As shown in Section~\ref{sec:trainingppl_res}, while improving task performance, DQO increases perplexity over the training data used during supervised training. This, combined with the finding that DQO is a suitable task alignment algorithm, is evidence for the existence of the task--data mismatch.

Much of this improvement can likely be credited to general, language-agnostic changes in model behavior, even with the restriction to using only 5 of the 30 supported target languages in DQO. If task alignment of a model with a given target language reduces the likelihood of untranslated source text, for instance, it would not be surprising to see similar improvements in other target languages.

Similarly, if task alignment for a given target language led to language-specific improvements (e.g., in grammar, sentence structure, punctuation, general fluency, etc.), it seems plausible that transfer learning could lead to improvements in closely related languages that have similar features.

However, manual inspection of translations before and after DQO revealed language-specific improvements in unrelated languages. In Latvian, for instance, foreign names are transliterated to match Latvian orthography and declined for grammatical case and gender, e.g. \citet{latvian_names_2021} report that \textit{George Clooney} should be translated as \textit{Džordžs Klūnijs}. While the baseline model applies correct transliteration occasionally and inconsistently, the DQO model almost always produces the correct transliteration. Several examples are included in Table~\ref{tab:latvian}.

\begin{table*}[t]
\small
\centering
\begin{tabular}{ll | rrrr | rrr | r}
\toprule
& & \multicolumn{4}{c |}{\textbf{Severity}} & \multicolumn{3}{c |}{\textbf{Language Specific}} & \\
\textbf{Language} & \textbf{Model} & \textbf{NT} & \textbf{Major} & \textbf{Minor} & \textbf{Trivial} & \textbf{Yes} & \textbf{No} & \textbf{N/A} & \textbf{Weighted MQM $\downarrow$}\\
\midrule
\multirow{2}{*}{Japanese}
 & Baseline & 0 & 1.15 & \textbf{0.61} & 0.06 & 1.28 & 0.50 & 0.01 & 6.256 \\
 & DQO & 0 & \textbf{0.93} & 0.63 & \textbf{0.03} & \textbf{1.16} & \textbf{0.40} & 0.01 & \textbf{5.223} \\
\midrule
\multirow{2}{*}{Lithuanian}
    & Baseline & 0.03 & 0.95 & 0.89 & 0.12 & 1.48 & 0.51 & 0 & 6.402 \\
    & DQO  & \textbf{0.01} & \textbf{0.80} & \textbf{0.77} & \textbf{0.10} & \textbf{1.24} & \textbf{0.44} & 0 & \textbf{5.030} \\
\bottomrule
\end{tabular}
\caption{\textbf{Mean number of Multidimensional Quality Metrics (MQM) errors per segment}, as annotated by professional human evaluators, with two different groupings: by severity and by whether the MQM subcategory is language specific or agnostic. NT stands for non-translation, i.e., a segment that cannot be construed as a translation of the source. Trivial refers to minor punctuation errors. This covers 100 randomly sampled English segments from the FLORES+ dataset, translated by the NVIDIA Megatron model before task alignment (baseline) and after task alignment (DQO). The weighted MQM score follows \citet{freitag-etal-2021-experts}.}
\label{tab:mqm_scores}
\end{table*}

As DQO was only performed on Chinese, German, Hindi, Russian or Spanish, none of which are closely related to Latvian, this behavior cannot have been learned from scratch during DQO. Although Chinese, Hindi, and Russian also transcribe foreign names, they use non-Latin scripts.

One possible explanation is that the baseline model learned to model both transliteration and non-transliteration, due to the range of translation quality in its supervised training data, causing inconsistent behavior at inference time. When DQO then shifts the output distribution towards certain human-preferred features, the probability of any correlated features (e.g., transliteration in Latvian), also increases.

\subsection{Human Evaluation}

To verify the presence of further language-specific changes for unrelated languages, we performed a human evaluation using the Multidimensional Quality Metrics framework (MQM) with professional translators \citep{mqmframework2014, freitag-etal-2021-experts}. The translators were trained on MQM and Anthea\footnote{\url{https://github.com/google-research/google-research/tree/a676d87/anthea}}, the open-source tool we used for performing MQM. We follow \citet{freitag-etal-2021-experts} in weighting major non-translations at 25 MQM points, other major errors at 5, and all minor errors at 1, except minor punctuation errors, which are 0.1 points.

For analysis, we selected two target languages not closely related to the languages used for task alignment: Lithuanian and Japanese.

These were selected to provide one low-to-medium resource language written in the Latin script and one in a non-Latin script, because neither is an outlier in quality metric improvement compared to the other supported language pairs, and to avoid the bias of examining Latvian, which we had already manually inspected.

For each language, we sampled complete documents (each generally two to five sentences forming a single paragraph) from FLORES+ until we had 100 source segments. The translators then annotated the baseline and task-aligned translations.

We then sorted the MQM error subcategories into two buckets, language agnostic and language specific, as seen in Table~\ref{tab:mqm_error_types} in Appendix~\ref{appendix:mqm}.

We observe reduced error rates in both Japanese and Lithuanian in both the language-agnostic and language-specific categories (Table~\ref{tab:mqm_scores}). The overall weighted MQM score also decreased for both languages, with significant improvements in both Lithuanian ($p_u = .001$) and Japanese ($p_u = .012$), where $p_u$-values are conservative estimates of the true $p$-values computed using paired one-sided approximate randomization \citep{phipsonsmyth2010pvalues} with the Marot toolkit.\footnote{\url{https://github.com/google-research/google-research/tree/a676d87/marot/README.md}}

\begin{table*}[t]
\vskip 0.1in
\centering
\resizebox{1\linewidth}{!}{
\begin{tabular}{lcccccccc}
\toprule
& \multicolumn{4}{c}{\textbf{English}~$\rightarrow$~\textbf{Czech}} & \multicolumn{4}{c}{\textbf{English}~$\rightarrow$~\textbf{German}}\\
\cmidrule(lr){2-5} \cmidrule(lr){6-9}
\multirow{-2}{*}{\textbf{ Model}} & BLEURT & COMET22 & CometKiwi22 & BLEU & BLEURT & COMET22 & CometKiwi22 & BLEU \\
\midrule
ALMA-13B-LoRA      & 79.62          & 88.94          & 83.31          & \textbf{29.33} & 75.06          & 85.14          & 82.19          & 29.65 \\
+ DQO              & 80.58          & 89.69          & \textbf{84.46} & 27.72          & 76.03          & 85.95          & \textbf{83.10} & \textbf{29.72} \\
+ CPO (ALMA-13B-R) & \textbf{80.90} & \textbf{89.73} & 84.38          & 24.29          & \textbf{76.79} & \textbf{86.24} & 82.96          & 26.72 \\
\toprule
& \multicolumn{4}{c}{\textbf{English}~$\rightarrow$~\textbf{Icelandic}} & \multicolumn{4}{c}{\textbf{English}~$\rightarrow$~\textbf{Russian}}\\
\cmidrule(lr){2-5} \cmidrule(lr){6-9}
\multirow{-2}{*}{\textbf{ Model}} & BLEURT & COMET22 & CometKiwi22 & BLEU & BLEURT & COMET22 & CometKiwi22 & BLEU \\
\midrule
ALMA-13B-LoRA      & 71.64          & 85.32          & 80.84          & 25.06          & 74.25          & 86.90          & 82.55          & \textbf{27.48} \\
+ DQO              & \textbf{72.00} & 85.57          & \textbf{81.72} & \textbf{25.09} & 75.40          & 87.71          & \textbf{83.68} & 26.68 \\
+ CPO (ALMA-13B-R) & 71.71          & \textbf{86.25} & 81.20          & 21.03          & \textbf{75.74} & \textbf{88.05} & 83.63          & 23.12 \\
\toprule
& \multicolumn{4}{c}{\textbf{English}~$\rightarrow$~\textbf{Chinese (simpl.)}} & \multicolumn{4}{c}{\textbf{Average}}\\
\cmidrule(lr){2-5} \cmidrule(lr){6-9}
\multirow{-2}{*}{\textbf{ Model}} & BLEURT & COMET22 & CometKiwi22 & BLEU & BLEURT & COMET22 & CometKiwi22 & BLEU\\
\midrule
ALMA-13B-LoRA      & 69.79          & 85.54          & 80.56          & \textbf{37.80} & 74.07          & 86.37          & 81.89          & \textbf{29.86} \\
+ DQO              & \textbf{70.60} & \textbf{86.37} & \textbf{81.84} & 35.58          & 74.92          & 87.06 & \textbf{82.96} & 28.96 \\
+ CPO (ALMA-13B-R) & \textbf{70.60} & 86.35          & 81.79          & 32.15          & \textbf{75.15} & \textbf{87.32}          & 82.79          & 25.46 \\
\bottomrule
\end{tabular}
}
\caption{Evaluation of DQO and CPO \citep{xu2024cpo} on the ALMA-13B-LoRA model. Scores are reported on the WMT'21 (Icelandic) and WMT'22 (remaining languages) test sets. The hyperparameters are specified in Appendix~\ref{appendix:alma}.
}
\label{tab:alma_13b_lora}
\end{table*}

\subsection{DQO for Large Language Models}
To compare DQO's performance against the strong baseline of other state-of-the-art DPO variants on a large language model trained specifically for translation, we apply it to the Alma-13B-LoRA model, a LLaMA-2-13B model with continued pretraining on Chinese, Czech, English, German, Icelandic, and Russian monolingual data and LoRA fine-tuning on high quality translation data \citep{xu2024paradigmshiftmachinetranslation, hu2022lora}.

The highest performing human preference alignment method previously reported for this model is Contrastive Preference Optimization (CPO), a variant of DPO applied to the Alma-13B-LoRA model to create Alma-13B-R \citep{xu2024cpo}. To ensure a direct comparison of optimization methods, we adopt the same data conditions and parameter masks as that prior work: restricting our seed dataset to the training data used for Alma-13B-R (the combined FLORES+ dev and devtest splits), fine-tuning only the LoRA adapters of the model, and evaluating translation out of English on the WMT'21 (for Icelandic) and WMT'22 (for the other languages) datasets.

Due to the restricted seed dataset used in this experiment, source segments are reused between rounds. As in previous experiments, we sample 8000 source segments, sample 64 translations per segment (as well as the greedy translation), and use CometKiwi22 as a proxy for human preferences. Other hyperparameters were adjusted based on a manual hyperparameter search to accommodate the differing training and sampling dynamics of LoRA training with an LLM (see Appendix~\ref{appendix:alma} for all hyperparameters).

Table~\ref{tab:alma_13b_lora} shows the results. The translations for ALMA-13-LoRA and ALMA-13B-R are generated with greedy inference on the publicly available model parameters\footnote{\url{https://huggingface.co/haoranxu/ALMA-13B-Pretrain-LoRA}, \url{https://huggingface.co/haoranxu/ALMA-13B-R}}. This experiment indicates that DQO maintains a substantially higher BLEU score than CPO while providing similar improvements in BLEURT, COMET22, and CometKiwi22. Unlike our encoder-decoder experiments, source segments were reused between rounds to achieve a fair comparison with CPO. We would expect a higher performance with a larger pool of source data, but leave confirmation of this assumption to future work.

\section{Related Work}
The idea of task--data mismatch in NMT is not new. There has been extensive previous work focused on reducing this mismatch through data filtering, using surface-level heuristics \cite{koehn-etal-2007-moses}, statistical and neural models for alignment and quality evaluation \cite{prompsit:2018:WMT, heffernan-etal-2022-bitext, peter-etal-2023-theres}, language identification \cite{lui-baldwin-2011-cross, joulin2016bag}, or ensembles \cite{koehn-etal-2020-findings}.

While data filtering techniques do help reduce the task--data mismatch, they force a trade-off between increasing task alignment and retaining flawed, but potentially useful, training data. To counter this, curriculum learning can be used, by training first on a conservatively filtered dataset, then shifting to a cleaner subset of the data \cite{bogoychev2023opuscleaneropustraineropensource}.

However, no amount of data filtering can remove the effects of translationese, as it is present in all translations. \citet{riley-etal-2020-translationese} and \citet{freitag-etal-2022-natural} both address this by treating original and translated text as separate languages in a "multilingual" NMT model, by training either a classifier or a contrastive language model to tag each source and target segment as either original or translated. At inference time, they use their model in a zero-shot setting to translate from original source text into the distribution of original target text.

Similarly, \citet{tomani2024qualityawaretranslationmodelsefficient} label each source sentence with a binned QE score. By adding the label of the highest quality bin to a source sentence at inference time, they successfully bias the model towards high quality translations.

\citet{ramos2024aligningneuralmachinetranslation} apply RLHF \citep{ziegler2020rlhf} to NMT using various QE metrics as reward, and compare it to data filtering, re-ranking using a QE model, and Minimum Bayes Risk decoding (MBR) \cite{kumar-byrne-2004-minimum, freitag2022highquality}, finding that a combination of data filtering, RLHF, and re-ranking performs best.

In DPO MBR fine-tuning, MBR is used to generate preference pairs for use with DPO \cite{yang-etal-2024-direct}. Compared to DQO, this method is computationally more expensive, and requires a reference-based QE model. In addition, DQO's online nature ensures that preference pairs remain relevant to the policy model.

\citet{xu2024advancingtranslationpreferencemodeling} apply RLHF with a reward model trained to distinguish high quality references (from literary translations) and translations sampled from their model. Similar to us, they find evidence of cross-lingual transfer learning during preference learning. Specifically, when optimized only on EN--ZH, their model improved for EN to FR, ES, RU, and AR. When training only on EN--AR, however, they saw improvements in only half of the target languages.

Reward rAnked Fine-Tuning (RAFT) is the method most similar to DQO, but uses SFT to update the model towards a single preferred output rather than using DPO with a preferred/rejected output pair \cite{dong2023raft}. As it was not evaluated for the translation task, used an independently trained reward model, and had slight differences in sampling parameters, we ran an ablation on whether to use DPO or SFT in DQO (see Section~\ref{sec:DQO}).

\section{Conclusion}
We demonstrate the existence of a fundamental task--data mismatch in NMT and introduce Direct Quality Optimization (DQO), a method of aligning pretrained models with human preference.

Using DQO on a multilingual NMT model, we find improvements in automatic quality metrics for all supported target languages, even those neither used for DQO, nor related to the languages used for DQO.  A human evaluation confirms that these improvements reflect increased human preference.

The improvements in translation quality for unrelated languages include language specific features that were not seen during DQO, suggesting that the baseline model had, but did not use, knowledge of those features during inference. We suggest that this is the expected behavior of a model trained with supervised learning, and present DQO as an efficient method of aligning a translation model with human preference.

In an experiment on ALMA-13B-LoRA we confirm that DQO is applicable to decoder-only LLMs.

\section{Limitations}
This work only tests one quality evaluation model as a proxy for human preferences, CometKiwi22, and does not examine the impact of that proxy's quality. We focused primarily on a single translation model, the NVIDIA Megatron English-Many model, using a 1.3B paramter English-German model only for the perplexity experiments (as we had access to the training data), and ALMA-13B-LoRA to verify applicability on decoder-only models. Human evaluation of translation quality was only performed on two language pairs. For all others, we relied on automatic quality evaluation metrics such as BLEURT, COMET22 and BLEU, which may not fully capture true human preference.

\bibliography{anthology,custom}

\clearpage
\appendix
\onecolumn
\section{Appendix}

\subsection{MQM Error Subcategories by Generality}
\label{appendix:mqm}

\begin{table*}[h]
\centering
\begin{tabular}{lll}
\toprule
\textbf{Language-agnostic} & \textbf{Language-specific} & \textbf{Other} \\
\midrule
Accuracy/Creative Reinterpretation & Fluency/Grammar & Other \\
Accuracy/Mistranslation & Fluency/Register & Source issue \\
Accuracy/Source language fragment & Fluency/Spelling \\
Accuracy/Addition & Fluency/Punctuation \\
Accuracy/Omission & Fluency/Character encoding \\
Fluency/Inconsistency & Style/Unnatural or awkward \\
Terminology/Inconsistent & Style/Bad sentence structure \\
Non-translation & Terminology/Inappropriate for context \\
 & Locale convention/Address format \\
 & Locale convention/Date format \\
 & Locale convention/Currency format \\
 & Locale convention/Telephone format \\
 & Locale convention/Time format \\
 & Locale convention/Name format \\
\bottomrule
\end{tabular}
\caption{\textbf{Multidimensional Quality Metrics error subcategories by generality}. \textit{Language-agnostic errors} are those governed by a principle that can be generalized to all language pairs, e.g., that translations should not omit information. \textit{Language-specific errors} are those that require additional, language-specific information to generalize from one language pair to another, e.g., correcting improper sentence structure requires knowledge of correct vs.\ incorrect sentence structures for a given language. \textit{Other errors} cannot be assigned to either category.}
\label{tab:mqm_error_types}
\end{table*}

\subsection{Hyperparameters Used in Experiments on NVIDIA Megatron}
\begin{table*}[h]
\centering
\begin{tabular}{l l r}
\toprule
    Hyperparameter & Definition & Value \\
\midrule
    $r_{QE}$ & Human preference proxy model & \texttt{CometKiwi22} \\
    $n$ & Number of rounds & $5$ \\
    $m$ & Epochs per round & $8$ \\
    $d$ & Epoch size (source sentences) & $8000$ \\
    $\alpha$ & Learning rate & $1 \times 10^{-6}$ \\
    $\beta$ & DPO regularization factor & 0.5 \\
    $k$ & Sampled translations per source & 64 \\
    $K$ & Top-K sampling parameter & 40 \\
    $P$ & Top-P sampling parameter & 0.8 \\
    $\varepsilon$ & Preference margin & 0.005 \\
    -- & Batch size & 8096 \\
    -- & Learning rate schedule & Linear with warmup \\
    -- & Learning rate warmup steps & 150 \\
    -- & Gradient clipping threshold (norm) & 10 \\
\bottomrule
\end{tabular}
\caption{\textbf{A list of all hyperparameters} used for Direct Quality Optimization in this paper's experiments.}
\label{appendix:hyperparameters}
\end{table*}

\clearpage
\subsection{Hyperparameters for experiments on ALMA-13B-LoRA}
\label{appendix:alma}
\begin{table*}[h]
\centering
\begin{tabular}{l l r}
\toprule
    Hyperparameter & Definition & Value \\
\midrule
    $r_{QE}$ & Human preference proxy model & \texttt{CometKiwi22} \\
    $n$ & Number of rounds & $9$ \\
    $m$ & Epochs per round & $4$ \\
    $d$ & Epoch size (source sentences) & $8000$ \\
    $\alpha$ & Learning rate & $5 \times 10^{-5}$ \\
    $\beta$ & DPO regularization factor & 0.5 \\
    $k$ & Sampled translations per source & 64 \\
    $K$ & Top-K sampling parameter & $\infty$ \\
    $P$ & Top-P sampling parameter & 1.0 \\
    $\varepsilon$ & Preference margin & 0.005 \\
    -- & Batch size & 8096 \\
    -- & Learning rate schedule & Linear with warmup \\
    -- & Learning rate warmup steps & 150 \\
    -- & Gradient clipping threshold (norm) & 10 \\
\bottomrule
\end{tabular}
\caption{\textbf{A list of all hyperparameters} used for Direct Quality Optimization in the experiments on ALMA-13B-LoRA. See 
\url{https://github.com/lilt/dqo/blob/main/configs/alma-13b-lora-comparison-with-cpo-4.yaml}}
\label{appendix:hyperparameters-alma-13b}
\end{table*}

\subsection{Composition of the DQO Seed Dataset}\label{appendix:seed_dataset}
As described in Figure \ref{alg:DQO}, Direct Quality Optimization requires a seed dataset containing input samples in the source language. This dataset does not need to include references, as the policy model $\pi_\theta$ is used to produce a diverse set of hypotheses, which are then scored under a QE model and transformed into preference pairs.

For our experiments, we used a general and varied seed dataset consisting of the English side of the following publicly available English--German datasets provided by the OPUS project \cite{tiedemann-2012-parallel}:
\begin{itemize}
    \item bible-uedin \cite{christodouloupoulos2015}
    \item CCAligned \cite{el-kishky-etal-2020-ccaligned}
    \item CCMatrix \cite{schwenk-etal-2021-ccmatrix, fanetal2021}
    \item DGT v2019\footnote{\url{https://ec.europa.eu/jrc/en/language-technologies/dgt-translation-memory}. The European Commission retains ownership of the data.}
    \item EBC
    \item ELRA-W0143\footnote{\url{https://www.elrc-share.eu}}
    \item ELRA-W0201
    \item ELRC-CORDIS\_News\footnote{\url{https://elrc-share.eu/repository/browse/english-french-parallel-corpus-from-cordis-project-news/e4597da00ae511e9b7d400155d026706c248250ecee54d19bef388d2a42e6d93/}}
    \item ELRC-CORDIS\_Results\footnote{\url{https://elrc-share.eu/repository/browse/german-english-parallel-corpus-from-cordis-project-results-in-brief/e70e0b920ae511e9b7d400155d026706b079d7cd7f984a98ab96380f6215f358/}}
    \item ELRC-EMEA\footnote{\url{https://elrc-share.eu/repository/browse/bilingual-corpus-made-out-of-pdf-documents-from-the-european-medicines-agency-emea-httpswwwemaeuropaeu-february-2020-en-de/d6ce198a862611ea913100155d0267064011b731322946a6b897cf495fb6f023/}. This dataset has been generated out of public content available through European Medicines Agency: \url{https://www.ema.europa.eu/}, in February 2020.}
    \item ELRC-EU\_publications\footnote{This dataset was generated from public content available through the Publications Office of the European Union (OP Portal), \url{https://op.europa.eu/en/home}}
    \item ELRC-EUR\_LEX\footnote{\url{https://elrc-share.eu/repository/browse/covid-19-eur-lex-dataset-ilingual-en-mt/cf57fe82c5af11ea913100155d026706b5596d3f449a456f983bbb4e23de81a4/}}
    \item ELRC-Information\_Portal\footnote{\url{https://elrc-share.eu/repository/browse/information-portal-of-the-czech-president-and-czech-castle/2c11868e088b11e6b68800155d020502c402eaf049834da0bbb019049e42098c/}}
    \item ELRC-presscorner\_covid\footnote{\url{https://elrc-share.eu/repository/browse/covid-19-eu-presscorner-v1-dataset-bilingual-en-de/67c1519c969311ea913100155d0267063c11069dcb104114901b3160c9f7618c/}}
    \item EMEA
    \item EUBookshop
    \item EUConst
    \item EuroPat\footnote{\url{https://europat.net/}}
    \item GlobalVoices
    \item GNOME
    \item JRC-Acquis v3.0 \cite{jrcacquis2006}\footnote{\url{https://joint-research-centre.ec.europa.eu/language-technology-resources/jrc-acquis_en}. The European Commission retains ownership of the data.}
    \item KDE4
    \item LinguaTools-WikiTitles
    \item MultiUN \cite{eisele-chen-2010-multiun}
    \item News-Commentary \cite{kocmi-etal-2023-findings}
    \item OpenSubtitles \cite{lison-tiedemann-2016-opensubtitles2016}
    \item ParaCrawl \cite{banon-etal-2020-paracrawl}
    \item PHP
    \item Tatoeba
    \item Tilde EESC \cite{rozis-skadins-2017-tilde}
    \item TildeMODEL \cite{rozis-skadins-2017-tilde}
    \item WikiMatrix \cite{schwenk-etal-2021-wikimatrix}
    \item wikimedia\footnote{\url{https://dumps.wikimedia.org/other/contenttranslation/}}
    \item Wikipedia \cite{WOLK2014126}
    \item Wikititles \cite{kocmi-etal-2023-findings}
    \item XLEnt \cite{el2021xlent}
\end{itemize}

As well as the following publicly available datasets which were not obtained through OPUS:
\begin{itemize}
    \item ELITR ECA \cite{williams2021elitrecacorpus}
    \item Europarl \cite{koehn-2005-europarl}
    \item Tilde EMA \cite{rozis-skadins-2017-tilde}
    \item Tilde RAPID 2019 \cite{rozis-skadins-2017-tilde}
    \item WIPO COPPA \cite{coppa2016}
    \item WMT13 CommonCrawl \cite{smith-etal-2013-dirt}
\end{itemize}

These datasets were also used to train the model used in Section~\ref{sec:trainingppl_res}.
\clearpage

\subsection{Results by Target Language}\label{appendix:evalmetrics}
\begin{table*}[hbt!]
    \centering
    \begin{tiny}
    \begin{tabular}{llrrrrrrrr}
    \toprule
    \multirow{2}{*}{\small{Model}} & \multirow{2}{*}{\small{Lang.}} & \multicolumn{4}{c}{\small{FLORES+ devtest}} & \multicolumn{4}{c}{\small{NTREX}} \\
     &  & \small{BLEURT} & \small{COMET22} & \small{CometKiwi22} & \small{BLEU} & \small{BLEURT} & \small{COMET22} & \small{CometKiwi22} & \small{BLEU} \\
    \midrule
    Baseline & bg & 0.8400 & 0.8974 & 0.8524 & 41.80 & 0.7713 & 0.8520 & 0.8242 & 32.00 \\
    DQO & bg & \textbf{0.8526} & \textbf{0.9067} & \textbf{0.8614} & \textbf{42.70} & \textbf{0.7865} & \textbf{0.8638} & \textbf{0.8341} & \textbf{32.40} \\
    \midrule
    Baseline & cs & 0.7758 & 0.8826 & 0.8327 & 32.60 & 0.7282 & 0.8509 & 0.8065 & 30.10 \\
    DQO & cs & \textbf{0.7978} & \textbf{0.9002} & \textbf{0.8504} & \textbf{34.00} & \textbf{0.7506} & \textbf{0.8696} & \textbf{0.8255} & \textbf{30.70} \\
    \midrule
    Baseline & da & 0.7744 & 0.8942 & 0.8396 & 46.40 & 0.7136 & 0.8541 & 0.8145 & 37.40 \\
    DQO & da & \textbf{0.7948} & \textbf{0.9091} & \textbf{0.8565} & \textbf{48.60} & \textbf{0.7355} & \textbf{0.8721} & \textbf{0.8341} & \textbf{39.30} \\
    \midrule
    Baseline & de & 0.7417 & 0.8535 & 0.8222 & 38.80 & 0.6793 & 0.8100 & 0.7950 & 30.80 \\
    DQO & de & \textbf{0.7561} & \textbf{0.8682} & \textbf{0.8338} & \textbf{39.30} & \textbf{0.7041} & \textbf{0.8315} & \textbf{0.8117} & \textbf{31.80} \\
    \midrule
    Baseline & el & 0.6738 & 0.8641 & 0.8032 & 25.90 & 0.6477 & 0.8494 & 0.7876 & 30.60 \\
    DQO & el & \textbf{0.6793} & \textbf{0.8699} & \textbf{0.8044} & \textbf{26.60} & \textbf{0.6567} & \textbf{0.8585} & \textbf{0.7892} & \textbf{31.60} \\
    \midrule
    Baseline & es & 0.7467 & 0.8567 & 0.8569 & 27.50 & 0.7304 & 0.8474 & 0.8330 & 40.50 \\
    DQO & es & \textbf{0.7594} & \textbf{0.8656} & \textbf{0.8662} & \textbf{28.80} & \textbf{0.7421} & \textbf{0.8547} & \textbf{0.8425} & \textbf{41.00} \\
    \midrule
    Baseline & et & 0.7779 & 0.8792 & 0.8421 & 27.10 & 0.7279 & 0.8451 & 0.8155 & 24.20 \\
    DQO & et & \textbf{0.8114} & \textbf{0.9041} & \textbf{0.8647} & \textbf{28.90} & \textbf{0.7603} & \textbf{0.8690} & \textbf{0.8399} & \textbf{25.00} \\
    \midrule
    Baseline & fi & 0.7959 & 0.8899 & 0.8471 & 24.40 & 0.7393 & 0.8550 & 0.8247 & 18.70 \\
    DQO & fi & \textbf{0.8264} & \textbf{0.9105} & \textbf{0.8640} & \textbf{26.00} & \textbf{0.7640} & \textbf{0.8736} & \textbf{0.8421} & \textbf{19.60} \\
    \midrule
    Baseline & fr & 0.7400 & 0.8638 & 0.8486 & 49.40 & 0.6525 & 0.8221 & 0.8289 & 36.10 \\
    DQO & fr & \textbf{0.7529} & \textbf{0.8713} & \textbf{0.8544} & \textbf{50.70} & \textbf{0.6632} & \textbf{0.8305} & \textbf{0.8344} & \textbf{37.00} \\
    \midrule
    Baseline & hi & 0.6825 & 0.7645 & 0.8040 & \textbf{32.90} & 0.6313 & 0.7227 & 0.7735 & \textbf{25.50} \\
    DQO & hi & \textbf{0.6991} & \textbf{0.7862} & \textbf{0.8217} & 32.50 & \textbf{0.6511} & \textbf{0.7459} & \textbf{0.7972} & 25.10 \\
    \midrule
    Baseline & hr & 0.8190 & 0.8942 & 0.8624 & 31.10 & 0.7707 & 0.8644 & 0.8326 & 31.80 \\
    DQO & hr & \textbf{0.8318} & \textbf{0.9032} & \textbf{0.8695} & \textbf{32.10} & \textbf{0.7847} & \textbf{0.8770} & \textbf{0.8445} & \textbf{32.50} \\
    \midrule
    Baseline & hu & 0.8378 & 0.8645 & 0.8354 & 26.90 & 0.7616 & 0.8141 & 0.8118 & 17.40 \\
    DQO & hu & \textbf{0.8554} & \textbf{0.8800} & \textbf{0.8488} & \textbf{27.10} & \textbf{0.7793} & \textbf{0.8294} & \textbf{0.8268} & \textbf{18.00} \\
    \midrule
    Baseline & id & 0.8030 & 0.9092 & 0.8414 & 47.50 & 0.7648 & 0.8823 & 0.8111 & 40.50 \\
    DQO & id & \textbf{0.8158} & \textbf{0.9172} & \textbf{0.8516} & \textbf{49.30} & \textbf{0.7784} & \textbf{0.8917} & \textbf{0.8251} & \textbf{41.10} \\
    \midrule
    Baseline & it & 0.7699 & 0.8725 & 0.8590 & 30.60 & 0.7280 & 0.8455 & 0.8279 & 36.70 \\
    DQO & it & \textbf{0.7860} & \textbf{0.8821} & \textbf{0.8676} & \textbf{31.40} & \textbf{0.7467} & \textbf{0.8613} & \textbf{0.8434} & \textbf{37.50} \\
    \midrule
    Baseline & ja & 0.6832 & 0.8918 & 0.8545 & 32.60 & 0.6042 & 0.8584 & 0.8251 & 26.40 \\
    DQO & ja & \textbf{0.6981} & \textbf{0.9019} & \textbf{0.8629} & \textbf{34.10} & \textbf{0.6208} & \textbf{0.8713} & \textbf{0.8395} & \textbf{27.10} \\
    \midrule
    Baseline & ko & 0.6538 & 0.8689 & 0.8433 & 29.40 & 0.5788 & 0.8317 & 0.8085 & 25.50 \\
    DQO & ko & \textbf{0.6734} & \textbf{0.8820} & \textbf{0.8550} & \textbf{30.30} & \textbf{0.5980} & \textbf{0.8481} & \textbf{0.8250} & \textbf{26.50} \\
    \midrule
    Baseline & lt & 0.8043 & 0.8742 & 0.8344 & 27.30 & 0.7485 & 0.8404 & 0.8057 & 21.60 \\
    DQO & lt & \textbf{0.8264} & \textbf{0.8910} & \textbf{0.8490} & \textbf{28.80} & \textbf{0.7699} & \textbf{0.8564} & \textbf{0.8181} & \textbf{22.30} \\
    \midrule
    Baseline & lv & 0.7896 & 0.8677 & 0.8253 & 30.50 & 0.6997 & 0.8097 & 0.7816 & 20.40 \\
    DQO & lv & \textbf{0.8201} & \textbf{0.8902} & \textbf{0.8431} & \textbf{32.10} & \textbf{0.7418} & \textbf{0.8424} & \textbf{0.8088} & \textbf{21.70} \\
    \midrule
    Baseline & nl & 0.7425 & 0.8617 & 0.8483 & 27.00 & 0.7080 & 0.8384 & 0.8205 & 34.20 \\
    DQO & nl & \textbf{0.7611} & \textbf{0.8756} & \textbf{0.8601} & \textbf{28.10} & \textbf{0.7262} & \textbf{0.8556} & \textbf{0.8356} & \textbf{35.40} \\
    \midrule
    Baseline & no & 0.7771 & 0.8899 & 0.8526 & 33.80 & 0.7447 & 0.8622 & 0.8267 & 36.90 \\
    DQO & no & \textbf{0.7915} & \textbf{0.8991} & \textbf{0.8646} & \textbf{34.00} & \textbf{0.7644} & \textbf{0.8779} & \textbf{0.8445} & \textbf{38.70} \\
    \midrule
    Baseline & pl & 0.7600 & 0.8678 & 0.8206 & 21.40 & 0.6992 & 0.8312 & 0.7939 & 25.70 \\
    DQO & pl & \textbf{0.7787} & \textbf{0.8818} & \textbf{0.8312} & \textbf{22.80} & \textbf{0.7153} & \textbf{0.8463} & \textbf{0.8058} & \textbf{26.80} \\
    \midrule
    Baseline & pt & 0.7856 & 0.8941 & 0.8453 & 50.80 & 0.7069 & 0.8477 & 0.8236 & 33.90 \\
    DQO & pt & \textbf{0.7952} & \textbf{0.9000} & \textbf{0.8531} & \textbf{51.20} & \textbf{0.7197} & \textbf{0.8574} & \textbf{0.8341} & \textbf{35.00} \\
    \midrule
    Baseline & ro & 0.8026 & 0.8927 & 0.8594 & 40.30 & 0.7338 & 0.8441 & 0.8255 & 33.30 \\
    DQO & ro & \textbf{0.8144} & \textbf{0.9015} & \textbf{0.8645} & \textbf{41.40} & \textbf{0.7474} & \textbf{0.8571} & \textbf{0.8386} & \textbf{34.70} \\
    \midrule
    Baseline & ru & 0.7430 & 0.8755 & 0.8329 & 31.30 & 0.6706 & 0.8299 & 0.8002 & 31.80 \\
    DQO & ru & \textbf{0.7556} & \textbf{0.8842} & \textbf{0.8419} & \textbf{32.00} & \textbf{0.6831} & \textbf{0.8433} & \textbf{0.8104} & \textbf{31.90} \\
    \midrule
    Baseline & sl & 0.7978 & 0.8679 & 0.8359 & 30.00 & 0.7174 & 0.8106 & 0.7877 & 28.30 \\
    DQO & sl & \textbf{0.8252} & \textbf{0.8860} & \textbf{0.8517} & \textbf{31.80} & \textbf{0.7576} & \textbf{0.8410} & \textbf{0.8163} & \textbf{29.60} \\
    \midrule
    Baseline & sv & 0.7945 & 0.8957 & 0.8515 & 45.40 & 0.7401 & 0.8581 & 0.8192 & 40.90 \\
    DQO & sv & \textbf{0.8113} & \textbf{0.9064} & \textbf{0.8650} & \textbf{46.20} & \textbf{0.7632} & \textbf{0.8781} & \textbf{0.8400} & \textbf{42.40} \\
    \midrule
    Baseline & tr & 0.7693 & 0.8827 & 0.8441 & 29.10 & 0.6802 & 0.8235 & 0.8129 & 17.60 \\
    DQO & tr & \textbf{0.7875} & \textbf{0.8953} & \textbf{0.8559} & \textbf{30.10} & \textbf{0.7011} & \textbf{0.8402} & \textbf{0.8287} & \textbf{17.70} \\
    \midrule
    Baseline & uk & 0.7432 & 0.8728 & 0.8172 & 29.80 & 0.6678 & 0.8230 & 0.7838 & 24.80 \\
    DQO & uk & \textbf{0.7603} & \textbf{0.8878} & \textbf{0.8300} & \textbf{30.50} & \textbf{0.6868} & \textbf{0.8423} & \textbf{0.7983} & \textbf{25.80} \\
    \midrule
    Baseline & vi & 0.7157 & 0.8736 & 0.8299 & 42.20 & 0.6753 & 0.8442 & 0.8081 & 41.30 \\
    DQO & vi & \textbf{0.7329} & \textbf{0.8857} & \textbf{0.8429} & \textbf{43.80} & \textbf{0.6917} & \textbf{0.8589} & \textbf{0.8234} & \textbf{42.10} \\
    \midrule
    Baseline & zh & 0.7015 & 0.8582 & 0.8199 & 42.00 & 0.6267 & 0.8099 & 0.7879 & 34.50 \\
    DQO & zh & \textbf{0.7202} & \textbf{0.8752} & \textbf{0.8367} & \textbf{44.10} & \textbf{0.6468} & \textbf{0.8292} & \textbf{0.8067} & \textbf{36.00} \\
    \bottomrule
    \end{tabular}

    \end{tiny}
    \caption{Automatic quality evaluation metrics for all target languages supported by the NVIDIA Megatron model, before and after Direct Quality Optimization (DQO), computed on both the FLORES+ devtest and NTREX datasets. }
    \label{tab:flores_eval_all_langs}
\end{table*}

\subsection{Ablation of Update Step: DPO vs.\ SFT}
\label{appendix:dpo_vs_sft_ablation_flores_dev_all}

\begin{figure*}[hbt!]
    \centering
    \includesvg[width=0.8\textwidth]{dqo_ablation_dpo_vs_sft_flores_dev_langpairs.svg}
    \caption{\textbf{Mean BLEURT-20 per language pair on FLORES+ dev after each round of DQO} with the NVIDIA Megatron EN-X model, using either Direct Preference Optimization (DPO) or Supervised Fine-Tuning (SFT) to update the model. DQO with SFT is equivalent to Reward rAnked Fine-Tuning (RAFT).}
    \label{fig:dpo_vs_sft_ablation_flores_dev_all}
\end{figure*}

\begin{figure*}[hbt!]
    \centering
    \includesvg[width=0.95\textwidth]{dqo_ablation_dpo_vs_sft_flores_dev_avg_no_outliers.svg}
    \caption{\textbf{Mean BLEURT-20 on FLORES+ dev, excluding outliers after each round of DQO} with the NVIDIA Megatron EN-X model, using either Direct Preference Optimization (DPO) or Supervised Fine-Tuning (SFT) to update the model. DQO with SFT is equivalent to Reward rAnked Fine-Tuning (RAFT). English to French, Chinese, Japanese, and Korean were excluded from this chart as outliers. See Figure~\ref{fig:dpo_vs_sft_ablation_mean} for the chart including outliers.}
    \label{fig:dpo_vs_sft_ablation_flores_dev_avg_no_outliers}
\end{figure*}

\end{document}